\documentclass[conference]{IEEEtran}
\IEEEoverridecommandlockouts
\usepackage[T1]{fontenc}
\usepackage{textcomp}
\usepackage{amsmath,amssymb,amsthm,amstext}
\usepackage{newtxmath}
\usepackage{algorithm, algorithmic}
\usepackage{bbm}
\usepackage{color}
\usepackage{flushend}
\usepackage{graphicx}
\usepackage{wrapfig}
\usepackage{soul}
\usepackage[caption=false,font=footnotesize]{subfig}
\usepackage{rotating}
\usepackage{tikz}
\usetikzlibrary{automata,positioning}

\usepackage{mathtools}
\DeclarePairedDelimiter\floor{\lfloor}{\rfloor}

\usepackage{pgfplots} 
\pgfplotsset{compat=newest} 
\pgfplotsset{plot coordinates/math parser=false} 
\newlength\figureheight 
\newlength\figurewidth

\def\BibTeX{{\rm B\kern-.05em{\sc i\kern-.025em b}\kern-.08em
    T\kern-.1667em\lower.7ex\hbox{E}\kern-.125emX}}
\begin{document}

\title{Cross Layer Optimization and Distributed Reinforcement Learning for Wireless  \\ 360$^\circ$  Video Streaming}

\author{\IEEEauthorblockN{Anis Elgabli\IEEEauthorrefmark{1},
Mohammed S. Elbamby\IEEEauthorrefmark{2}, Cristina Perfecto\IEEEauthorrefmark{3}, Mounssif Krouka\IEEEauthorrefmark{4}, \\Mehdi Bennis\IEEEauthorrefmark{4},
and Vaneet Aggarwal\IEEEauthorrefmark{5}}
\IEEEauthorblockA{\IEEEauthorrefmark{1}Department of Industrial and Systems Engineering, Interdisciplinary Research Center for \\Communication Systems and
Sensing (IRC-CSS), Interdisciplinary Research Center for\\ Intelligent Secure Systems (IRC-ISS), King Fahd
University of Petroleum and Minerals (KFUPM), Dhahran, Saudi Arabia\\\IEEEauthorrefmark{2}Telefónica Research, Spain\\\IEEEauthorrefmark{3}Bilbao School of Engineering, University of the Basque Country (UPV/EHU), Spain\\\IEEEauthorrefmark{4}Centre for Wireless Communications, University of Oulu, Finland\\\IEEEauthorrefmark{5}School of Industrial Engineering, Purdue University, USA\\
Email: \IEEEauthorrefmark{1}anis.elgabli@kfupm.edu.sa,
\IEEEauthorrefmark{2}mohammed.elbamby@telefonica.com, \IEEEauthorrefmark{3}cristina.perfecto@ehu.eus,\\
\IEEEauthorrefmark{4}\{mounssif.krouka, mehdi.bennis\}@oulu.fi,
\IEEEauthorrefmark{5}vaneet@purdue.edu
}}

\maketitle

\begin{abstract}
Wirelessly streaming high quality 360 degree videos is still a challenging problem. When there are many users watching different 360 degree videos and competing for the computing and communication resources, the streaming algorithm at hand should maximize the average quality of experience (QoE) while guaranteeing a minimum rate for each user. In this paper, we propose a \emph{cross layer} optimization approach that maximizes the available rate to each user and efficiently uses it to maximize users' QoE. Particularly, we consider a tile based 360 degree video streaming, and we optimize a QoE metric that balances the tradeoff between maximizing each user's QoE and ensuring fairness among users. We show that the problem can be decoupled into two interrelated subproblems: (i) a physical layer subproblem whose objective is to find the download rate for each user, and (ii) an application layer subproblem whose objective is to use that rate to find a quality decision per tile such that the user's QoE is maximized. We prove that the physical layer subproblem can be solved optimally with low complexity and an actor-critic deep reinforcement learning (DRL) is proposed to leverage the parallel training of multiple independent agents and solve the application layer subproblem. Extensive experiments reveal the robustness of our scheme and demonstrate its significant performance improvement compared to several baseline algorithms.
\end{abstract}

\begin{IEEEkeywords}
360 video Wireless Streaming, Video Rate Adaptation, Virtual Reality (VR), Non Convex Optimization, Asynchronous advantage actor-critic (A3C), Deep Reinforcement Learning (DRL), millimeter wave (mmWave) communications.
\end{IEEEkeywords}
\section{Introduction}\label{sec:intro}  
Mobile video has emerged as a dominant contributor to the cellular traffic. It already accounts for more than half of the traffic and with new emerging technologies such as 360 degree videos, Virtual and Augmented Reality (VR and AR), it is estimated that the market could reach over \$130bn by 2029, with a compound annual growth rate of 42.05\% during that period (2024-2029)~\cite{vr_report}. 
To overcome the challenges of providing high quality video on-demand over cellular networks, development of rate adaptation techniques for video streaming gained significant attention in the past decade from both academia and industry \cite{FHuY}. 
Adaptation schemes can dynamically adjust the quality of the streamed video to the variations in network conditions through two main components, namely {\em content encoding} and {\em playback rate adaptation}.
	
Content encoding, performed at the server side, divides a video into chunks of fixed playback duration, at which each chunk is encoded at multiple resolutions that correspond to different bandwidth requirements. On the other hand, playback rate adaptation, which is implemented in the client, server, or in a middle box, can dynamically switch between the available quality levels during the playback time. The adaptation depends on many factors such as the available bandwidth or the client buffer occupancy. As a result, different chunks of the video could be displayed at different quality levels.
	
360 degree video streaming poses additional challenges to the streaming problem. For instance, the typical size of a 360 degree video chunk is significantly larger than the regular chunk size. Hence, streaming a 360 degree video requires massive amounts of data to be transmitted \cite{bastug2017interconnected}. Moreover, users typically watch 360 degree videos through virtual reality (VR) head-mounted displays (HMDs). However, the use of HMDs is very sensitive to motion-to-photon (MTP) latency, which needs to be kept below $20$ milliseconds~\cite{ElbambyVRMag18}. Reducing the latency calls for video transmission schemes that cut down the amount of data transmitted to only what the user actually sees on her field of view (FoV)~\cite{PerfectoVR2018}. For that, the idea of splitting the 360 degree chunk into tiles and delivering only the tiles that intersect with a user's FoV has been proposed in the literature \cite{ghosh2018robust, Qian2016tiling, ghosh2017rate}. However, wirelessly delivering the FoV portion of the 360 degree chunk in high quality and with tight latency bounds is still very demanding in terms of bandwidth consumption. Therefore, currently existing HMDs resort to either transmitting a low resolution content, or tethering using wires to provide the required high rate.
Recently, the use of millimeter wave frequency communications (mmWave) has been proposed as a high data rate wireless medium. MmWave is anticipated to be a major enabler for the fifth-generation (5G) bandwidth hungry applications such as high quality 360 degree video streaming. It promises huge spectrum bandwidth, but comes with implementation predicaments such as sensitivity to signal blockage and beam alignment complexity. Due to the high path loss in such high frequency, beamforming is necessary to form narrow directional beams towards the receiver/transmitter. These narrow beams increase the desired power towards the destination while decreasing the interference levels, at the cost of increasing the mmWave sensitivity to blockage. 
	
In this work we leverage mmWave communications and machine learning (ML) tools that allow a reinforcement learning (RL) agent to be trained in an offline manner to observe a network state, and perform an action that maximizes the system's reward to obtain the playback rate of the next chunk while maximizing the user's QoE.  
\subsection{Related Work}\label{sec:related}   
The predominant adaptive coding technique in use for conventional video streaming is adaptive video coding~(AVC~\cite{DASH}). In AVC, each video's chunk is encoded into multiple versions. Then, during the video downloading process, the player's adaptive bit rate (ABR) mechanism selects one of the versions based on its judgment of the network conditions and other aforementioned factors.

The conventional single user ABR problem has been extensively studied, with many streaming algorithms proposed. Researchers have investigated various approaches for streaming decisions, for example, by using the client's buffer state~\cite{BBA}, bandwidth prediction and buffer information~\cite{MPC,Miller15,elgabli2018fastscan}, Markov decision processes~\cite{Jarnikov11}, machine learning (ML)~\cite{Yang2022, mao2017neural, Kan2022}, and data-driven techniques~\cite{C3,CS2P}.	
ML tools, such as reinforcement learning (RL), have been commonly used in video streaming and in other resource allocation problems. Briefly, RL is defined by three components (state, action, and reward). For a given state, the RL agent is trained to choose the action that maximizes the cumulative discounted reward. The agent interacts with the environment in a continuous way and tries to find the best policy based on the reward/cost fed back from that environment \cite{ChenHZhang, ML2018}. 
For instance, authors in~\cite{mao2017neural} train an RL agent in an offline manner to solve a single user video streaming problem by observing some network states, and selecting the playback rate of the next chunk that maximizes the user's QoE.
The authors in~\cite{Zhang-360DRL} focus on jointly optimizing multiple QoE objectives through adaptively allocating rates for the tiles of the future video frames based on observations collected at the client video players. Similarly, in~\cite{MultiAgent-360DRL} a multi-agent DRL system is presented such that the users' QoE is maximized while minimizing the bandwidth consumption on the core network for multi-user live 360-degree video streaming. 

However, all the aforementioned works consider rate adaptation schemes that reside at the application layer of the client or the server. Hence, the client/server decides at what rate should the next chunk be downloaded based on the bandwidth prediction and/or the current buffer size. To improve the user's QoE in multi-user wireless environments where multiple users are competing for the resources, the physical layer resource management algorithm needs to take part in ensuring a minimum rate to each user (client). In this sense, QoE-aware resource allocation algorithms at the physical layer have been proposed in~\cite{elgabli2018qoe, cho2015qoe}, where the objective was to provide minimum rate guarantees to the ABR technique that works at the application layer, so that the ABR wisely use them to optimize the user's QoE. Indeed, a cross-layer approach that jointly considers QoE-aware physical resource allocation and the application layer ABR is sorely lacking. The latter, constitutes the main contribution of this paper.  

\subsection{Contribution}\label{sec:cont}   
In this paper, we tackle the problem of 360 degree video streaming in multi-user and mmWave wireless network scenario using a cross layer optimization approach. In particular, we formulate the problem as a cross layer optimization problem whose objective is to optimize a QoE metric that maintains a trade-off between maximizing the quality of each FoV tile in every chunk for every user, and ensuring fairness among the FoV tiles of each user, and also at the chunk and user level. The metric considers the concavity of the user's QoE with respect to the video playback rate~\cite{AnisSinglePath}, and the discrete nature of the video encoding rates.
	
The formulated problem is then decoupled into two subproblems which are solved online per chunk. The first is the physical layer subproblem whose objective is to maximize the download rates for each user while ensuring fairness. The second is the application layer subproblem whose objective is to use the achievable download rate of each user and distribute it among different tiles such that the user's QoE is maximized. We propose a low complexity algorithm to solve the physical layer subproblem, and show that even though the subproblem is non-convex, the proposed algorithm achieves the optimal solution under certain conditions. On the other hand, due to challenges posed by the application layer subproblem, such as the unavailability of non-causal information (future content) of the user's FoV, it is hard to find an optimal algorithm to solve it numerically. Therefore, we resort to a RL based method to undertake the application layer problem. Different from traditional RL approaches, we do not rely on a single RL agent, due to the resulting huge action space and computation overhead. In particular, for a single RL agent per user, the action space will exponentially grow with respect to the number of tiles and quality levels per tile. Alternatively, we consider a distributed RL approach in which an RL agent is considered for each tile. Each RL agent observes an input state and performs an action, while the reward is chosen such that the quality of each tile and the fairness among FoV tiles are both captured (see section~\ref{algo_appLayer}). With this scheme, each RL agent is trained to find the quality decision policy that maximizes the global reward. 
		
\begin{figure*}[t]
\centering
\includegraphics[width=0.9\textwidth,  height=5cm]{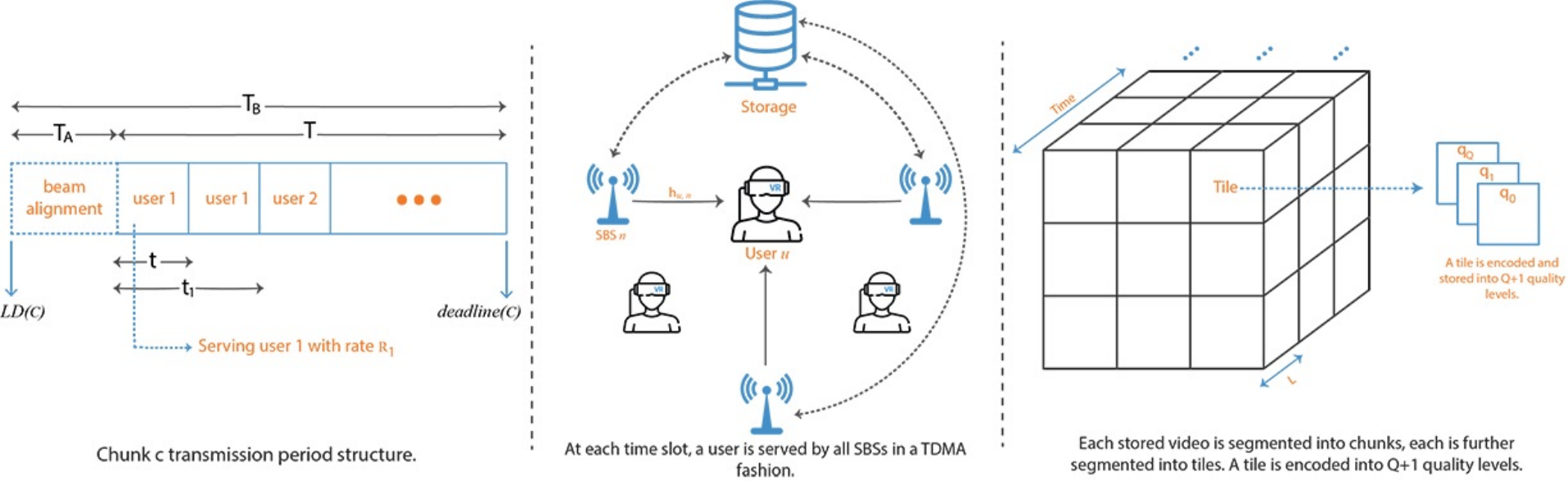}	
\caption{System model representing the transmission period structure (left), user scheduling through SBS cooperation (center), video segmentation into chunks and tiles logic (right).}
\label{fig:sysModel}
\end{figure*}  \vspace{-5pt}
\section{System Model}\label{sysModel}\vspace{-4pt}
Our system model and assumptions are described in Fig~\ref{fig:sysModel}. In this paper, we consider a system in which $N$ small base stations (SBS) with edge caching capabilities are serving a number of users watching different 360 degree videos. In particular, we assume that there are $U$ users, and each user $u \in \{1,\cdots,U\}$ is watching a 360 degree video $v \in \{1,\cdots,V\}$ divided into $C$ chunks, each chunk is of duration $L$ seconds. The video is played with an initial start-up, (i.e., buffering) delay $S$\footnote{We remark here that this assumption does not hinder or limit the scope of our problem formulation; As users will be independently scheduled from each other, the only implications of considering heterogeneous start-up times would be having a variable number of users actively competing for the resources at a given time slot.} and with a download deadline for each chunk where chunk $c$ needs to be downloaded by the time: $deadline(c)=S+(c-1)L-m$, i.e., the deadline of chunk $c$ is equal to the startup delay plus the time that is needed to play all chunks before $c$ minus a margin $m$. The margin $m$ is considered to ensure that the chunk is buffered some time before its playback time, so the possibility of having an empty buffer is minimized.  Each chunk is further divided into $J$ tiles, and each tile $j \in \{1,\cdots,J\}$ is encoded at one of $Q+1$ quality levels. To avoid users' competition over resources to download chunks they will not watch, and to have a better FoV prediction, we set a ``lower deadline'' for each chunk $c$ ($LD(c)$) such that the chunk cannot be downloaded before this lower dealine is met.

We further assume that the network operates in the mmWave band, and that at most one user is scheduled at each time slot $t$, with duration $\tau$. As shown in Fig~\ref{fig:sysModel}, the duration between $LD(c)$ and $deadline(c)$ is divided into two periods: (i) the beam alignment period ($T_A$) in which every SBS finds the best beam to serve each of the users, and (ii) the chunk download period $T$. Every user needs to download one chunk in the period $T$. Let $T_B$ be the beam coherence time which is assumed to be greater than the download time of one chunk, i.e., $T_B\geq T_a+T$. If a user $u$ is scheduled in time slot $t$ then, its achievable rate will be\vspace{-2pt}
\begin{equation}
R_u(t)=B\log_2\big(1+ \frac{P_u^t\sum_{n=1}^N|h_{u,n}f_{u,n}|^2}{N_oB}\big), \forall u,t
\label{equ:rateCon}
\end{equation}   
where $f_{u,n}$ is the directional beam gain of SBS $n$ toward user $u$ and $h_{u,n}$ is the channel between SBS $n$ and user $u$. We remark here that, as reflected by the summation inside the signal to noise ratio (SNR) term above, a coordinated transmission among SBSs as per single-frequency network (SFN) operation is assumed. The average rate achieved by user $u$ at the download period of each chunk is thus
\begin{equation}
\hat{R}_u= \frac{1}{T_B}\sum_{t=1}^{T}R_u(t), \forall u.
\label{equ:c1}
\end{equation} 

\section{Problem Formulation}\label{sec:problem}

In this section, we describe our problem formulation. Let the $q$-th quality level of tile $j$ in chunk $c$ for user $u$ be encoded at  rate $r_{u,c}^{j,q}$. Let ${X_{u,c}^{j,q}}=L*r_{u,c}^{j,q}$ be the size of tile $j$ in chunk $c$, when it is encoded at the $q$-th quality level. Note that a tile that is not fully downloaded by its respective deadline will not be displayed. Let us denote  $I_{u,c}^{j,q}$ the decision variable of the $q$-th level of tile $j$ in chunk $c$. $I_{u,c}^{j,q} \in \{0,1\}$. In other words, $I_{u,c}^{j,q}=1$ if tile $j$ of chunk $c$ is decided to be fetched at quality $q$, and $0$ otherwise. Moreover, we define the size difference between $q$ and $q-1$ quality levels $Y_{u,c}^{j,q}$ as follows:
	$$
	Y_{u,c}^{j,q} =    
	\begin{cases} 
		X_0 & q=1\\
		X_{u,c}^{j,q}- X_{u,c}^{j,q-1} & q>1.
	\end{cases}  
	$$
	
	Let $Z_{u,c}^{j,q}$ be the size that can be fetched out of $Y_{u,c}^{j,q}$. $Z_{u,c}^{j,q}=I_{u,c}^{j,q} * Y_{u,c}^{j,q} $. Since  $I_{u,c}^{j,q} \in \{0,1\}$, $Z_{u,c}^{j,q}$ is  $\in \{0,Y_{u,c}^{j,q}\}$. Further, let $z_{u,c}^{j,q}(t)$ be equal to the size fetched from $Z_{u,c}^{j,q}$ at time slot~$t$. 
	
Next, we explain the different metrics that we need to optimize before we introduce our optimization problem. In this work, we consider two metrics. The first metric is at the application layer, which can naturally be the average video quality in the FoV of each user. However, in order to ensure fairness among FoV tiles and among chunks, and to take into account the concave nature of user's QoE with respect to the video rate, we do not exactly consider the average video quality. Instead, we use a metric from \cite{elgabli2018fastscan}, which maximizes the quality of the chunk with the lowest quality. In our case, we consider the metric at the tile level in each user's FoV, in which maximizing the delivery of the tiles with lower qualities is prioritized. This metric is a weighted sum of $I_{u,c}^{j,q},\forall u,c,j,q$, in which the weights are chosen such that the delivery of the lower quality levels of all tiles is maximized, as will be explained later in this section. Hence, the application layer contribution to the objective function is defined as
	\begin{equation}
		\sum_{u=1}^{U}\sum_{c=1}^{C}\sum_{\substack{j=1,\\j\in\text{ Fov}}}^{J}\sum_{q=0}^{Q}(\beta^{q}I_{u,c}^{j,q}),
		\label{equ:AppObj}
	\end{equation}
where $\beta <1$. The choice of $\beta$ gives higher priority to fetching more FoV tiles at the $q^{th}$ quality level over fetching some at higher quality at the cost of dropping the quality of other tiles to below the $q^{th}$ quality level \cite{elgabli2018fastscan}.

The second metric that contributes to the objective function is a physical layer metric, namely the available rate to each user. Since users are watching videos encoded with different quality levels, in order to ensure fairness among users, we consider the ``{\it average FoV encoding rate violation}'' a metric which is defined for a user $u$, $w(u)\in \{{\cal W}: {\cal W}=w(u), \forall u \in \{1,\cdots,U\}\}$ as
	\begin{equation}
		w(u)=[R^{e}_u-{\hat R}_u]^+ \label{equ:phyObj}
	\end{equation}
where $[(\cdot)]^+=\text{max}\big((\cdot), 0\big)$ , and $R_u^{e}$ is the  ``average FoV encoding rate'' (AFER) of user $u$. The AFER is found as follows: given each tile is encoded in $\{0,1,\cdots,A-1\}$ quality levels, and the encoding rate of the quality level $j$ is $\zeta(j)$ Mbps, where $\zeta(0) < \zeta(1)<\cdots< \zeta(A-1)$, the average encoding rate of each tile ($a$) is defined as: $a=\frac{\sum_{i=0}^{A-1}\zeta(i)}{A}$ Mbps. Therefore, if there will be $K$ tiles in the FoV of a user at any given time, then, AFER, $R_{u}^{e}$, will be $K * a$ Mbps.
	
	%Therefore, if we define the rate available to user $u$ at time slot $t$ as $R_u(t)$, ouw metric will be:
	\if0
	\begin{equation}
		\Pr(R_u(t)<R^{e}_u)
		\label{equ:phyObj}
	\end{equation}
	\fi
	
Having defined the application and physical layer metrics to optimize, next we formulate the optimization problem that jointly \emph{(i)}  minimizes the maximum AFER among all users, and \emph{(ii)} maximizes the QoE metric described earlier. The optimization problem is formulated as
	
	\begin{subequations}
		\begin{align}
			 \textbf{Minimize}&~\!\Big(\!\!-\!\!\lambda\!\!\sum_{u=1}^{U}\!\sum_{c=1}^{C}\!\!\!\sum_{\substack{j=1,\\ j\in\text{ Fov}}}^{J}\!\!\!\sum_{q=0}^{Q}(\beta^{q}I_{u,c}^{j,q} )
			\!+\!(1\!-\!\lambda) \text{Max}(\cal{W})\!\Big)&\label{equ:mainObj}\\
			\vspace{-0.01cm}
		&\hspace{-0.75cm}\textrm{subject to} &\nonumber \\
			&\eqref{equ:rateCon}, \eqref{equ:phyObj}, \forall u,t
			\label{mainC0}\\
			& \sum_{c=1}^{C}\sum_{j=1}^{J}\sum_{q=0}^{Q}z_{u,c}^{j,q}(t) \leq R_{u}(t)\cdot \tau, \forall u,t  \label{equ:eq1c1}\\
			& I_{u,c}^{j,q} \leq I_{u,c}^{j,q-1}, \forall u,c,j,q>1
			\label{equ:eq1c3}\\
			& I_{u,c}^{j,q} \in \{0,1\}, \forall u,c,j,q
			\label{equ:eq1c4}\\
			& Z_{u,c}^{j,q}= I_{u,c}^{j,q}\cdot Y_{u,c}^{j,q},\quad  \forall u,c,j,q \label{equ:eq1c5}\\
			&\sum_{t=1}^{deadline(c)} z_{u,c}^{j,q}(t) = Z_{u,c}^{j,q},\forall u,c,j,q \label{equ:eq1c6}\\
			& z_{u,c}^{j,q}(t) \geq 0, \forall u,c,q,j,t \label{equ:eq1c7}\\
			& z_{u,c}^{j,q}(t)\! =\!\! 0, \forall u,c,q,j, t\!\!<\!\!LD(c)\!\text{ or }\! t\!\!>\!\!deadline(c) \label{equ:eq1c7_1}\\
			&\sum_{u=1}^{U}P_u^{t}=P, \forall t\label{equ:eq1c9}\\
			& P_u^{t}\in\{0,P\}, \forall u,t. \label{equ:eq1c10}
		\end{align}
	\end{subequations} 
	
In order to ensure  a low AFER, $\lambda \ll 1$ should be chosen. Constraint $\eqref{equ:eq1c1}$ forces every user to respect  its available rate at each time slot $t$. Constraint $\eqref{equ:eq1c3}$ ensures that a tile $j$ of chunk $c$ should not be considered for quality level $q$ if it is not a candidate to quality level $q-1$. Constraints $\eqref{equ:eq1c4}$ and $\eqref{equ:eq1c5}$ ensure that $Z_{u,c}^{j,q}$ can be either $0$ or $Y_{u,c}^{j,q}$, i.e, tile $j$ of chunk $c$ can either be downloaded at the $q$-th quality level or not. Constraint $\eqref{equ:eq1c6}$ defines $Z_{u,c}^{j,q}$ as the total amount that can be fetched for each tile of a chunk. Constraint $\eqref{equ:eq1c7}$ ensures the positivity of the download size at each time slot. Constraint $\eqref{equ:eq1c7_1}$ ensures that a chunk cannot be downloaded after its deadline. Finally, constraints $\eqref{equ:eq1c9}$ and $\eqref{equ:eq1c10}$ impose the maximum power constraint and ensure that only one user is assigned the physical resources at each time slot.

The problem defined in~\eqref{equ:mainObj}-\eqref{equ:eq1c10} is a non-convex optimization problem that has integer (non-convex) constraints. Integer-constrained problems are known to be NP hard in general \cite{nemhauser1988integer}. Very limited problems in this class of discrete optimization are known to be solvable in polynomial time. Moreover, the FoV of each user is not known non-casually. In addition, the available rate can be calculated only for a short time ahead (the beam coherence time). Therefore, we propose to solve the problem per chunk and approximate its solution by decoupling it into two subproblems. The objective of the first subproblem is to find the rate of each user $R_{u}(t)$ for the next beam coherence period such that $\text{Max}(\cal{W})$ is minimized, and the second subproblem is to maximize the application layer's QoE metrics given $R_{u}(t)$ of each user. Next, we discuss the two subproblems and then we propose our framework to solve the resulting optimization problem.
	
\subsection{Physical Layer Subproblem}\label{PhyAlgo}

In order to make the download rate decision of each user, the physical layer subproblem (PHY-LS) is solved before the download of every chunk. Note that a beam alignment period is considered before the download of each chunk, and that the beam coherence time is assumed to be longer than the chunk download time. Therefore, the PHY-LS is a time scheduling problem per beam coherence time, i.e, its goal is to find which user is scheduled to have the resources for each time slot for the upcoming beam coherence time. 
	
Our objective twofold: (i) minimize $\text{Max}(\cal{W})$ and, (ii) use the remaining resources to maximize the available rate for each user, so the application layer subproblem (APP-LS) can use the rate to maximize the QoE of each user. Since a user's QoE is a concave function with respect to the achievable rate \cite{elgabli2018fastscan}, and fairness among users also needs to be ensured, we consider the following optimization problem:

\begin{subequations}
\begin{align}
			\textbf{Minimize}& \Big(-\lambda\sum_{u=1}^U\big(\sum_{t=1}^{T}\log(R_u(t))\big)+(1-\lambda) {\text {Max}}({\cal W}) \Big)\label{equ:mainObjPhy}\\
			&\hspace{-0.5cm}\textrm{subject to}\nonumber \\ 
			&\hspace{0.5cm}\eqref{equ:rateCon}, \text{ } \eqref{equ:c1},\text{ } \eqref{equ:phyObj}, \forall u \label{phyC1}\\
			&\hspace{0.5cm}\hat{R}_u+ w_{u} \geq R_u^{e}, \forall u \label{phyC2}\\
			&\hspace{0.5cm}  w_{u}\geq 0, \forall u  \label{phyC3}\\
			&\hspace{0.5cm}(\ref{equ:eq1c9})-(\ref{equ:eq1c10}) \label{phyC4}
		\end{align}
	\end{subequations}
Note here that on choosing $\lambda \ll 1$ higher priority is given to avoiding falling bellow $R_{u}^{e}$ at any given time. Moreover, we maximize a sum log rate to ensure fairness among users. It is remarked that the choice of sum log rate serves the application layer objective which can be approximated more accurately by a continuous log function rather than by a linear function since higher quality levels contribute much less than lower ones to the total objective. Section~\ref{algo_appLayer} provides the details on our proposed algorithm to solve this problem for which it is shown that it achieves its optimal solution as $\lambda \rightarrow 0$. Next, the application layer subproblem is described.
\subsection{Application Layer Subproblem}\label{AppAlgo}
The PHY-LS described in the previous subsection finds which user is scheduled to use the resources during the time slots comprised in the next beam coherence interval. Hence, the achievable rate for each user during the next beam coherence period is exposed to the application layer algorithm which leverages this knowledge to download video tiles such that the application layer's QoE metric described in section~\eqref{equ:AppObj} is maximized. Therefore, the application layer optimization problem (APP-LS) can be solved per user since the users' problems are decoupled after finding the achievable rate for each one of them. Thus, for each user we have the following: \vspace{-0.4cm}
	\begin{subequations}
		\begin{align}
			\textbf{Maximize}&\,\, \sum_{c=1}^{C}\sum_{\substack{j=1,\\ j\in\text{ Fov}}}^{J}\sum_{q=0}^{Q}\beta^{q}I_{u,c}^{j,q} \label{perUserApp}\\
			&\hspace{-0.5cm}\textrm{subject to}\hspace{0.5cm}\eqref{mainC0}-\eqref{equ:eq1c7_1}.
		\end{align}
	\end{subequations}
	The above APP-LS is hard to solve since it is a non-convex with non-causal knowledge of the user's FoV. Moreover, the tiles are encoded with variable bitrate (VBR), so even tiles sharing the same quality level could have been encoded into different rates. Given all these metrics that contribute to the complexity of the problem, we propose an efficient algorithm based on machine learning (ML) tools. In particular, we use reinforcement learning (RL) to train our framework and solve the problem. We discuss our RL based algorithm in section~\ref{algo_appLayer}.

	\section{Proposed Algorithm to Solve the Physical Layer Subproblem}
	\label{phyAlgo}
	We now describe the proposed algorithm to solve the optimization problem \eqref{equ:mainObjPhy}-\eqref{phyC4}. The Algorithm is also outlined in Table \ref{table:bbm}. The first step of the algorithm is to index users randomly, and temporarily allocate a random amount of time slots to each user such that the sum of the allocated time slots is equal to $T$. For example, for $U$ users, one way is to allocate $\floor{\frac{T}{U}}$ time slots  to each user $\in\{1,\cdots,U\}$ and assign the remaining time slots to the $U$-th user. Note that this time slot assignment is not final since it may change when running the second phase of the algorithm.  As a result of the first step of the algorithm, user $u$ will be assigned $t_u$ time slots. Moreover, since the channel is constant in the beam coherence time, the achievable rate of user $u$ at any time slot in which the user is scheduled is the same, we have that 
	\begin{equation}
		r_u=R_u(t)=B\log(1+ \frac{P\sum_{n=1}^N|h_{u,n}f_{u,n}|^2}{N_oB})
		\label{equ:eq1c1Phy}
	\end{equation}
	With this, the average rate of this user is given by $R_u^{avg}=\frac{t_u}{T_B} \cdot r_u$ and the average FoV encoding rate violation  $w_u$ is
	\begin{equation}
		w_u =    
		\begin{cases} 
			R_u^{e} - \frac{t_u}{T_B}\cdot r_u & \text{ if } R_u^{e} > \frac{t_u}{T_B}\cdot r_u \\
			0 & \text{ otherwise.}
		\end{cases}  
		\label{violationEq}
	\end{equation}
		 \vspace{-0.4cm}

	The second step is to run an iterative algorithm, and at every iteration $k$, find the user whose index $u^\prime$ such that:
	\begin{equation}
		u^\prime=\underset{u}{\text {argmax}}(-\lambda t_{u}^{(k-1)}\log(r_{u})+(1-\lambda) w_{u}^{(k-1)})
		\label{c8Phy}
	\end{equation}
	Once $u^\prime$ is found, we choose another user $u^{\prime\prime}$ such that when:
	\begin{equation}
		t_{u^{\prime\prime}}^{(k)}=t_{u^{\prime\prime}}^{(k-1)}-1
		\label{c10Phy}
	\end{equation}
	and
	\begin{equation}
		w_{u^{\prime\prime}}^{(k)}=R_{u^{\prime\prime}}^{e}-\frac{t_{u^{\prime\prime}}^{(k)}}{T_B}\cdot r_{u^{\prime\prime}}
		\label{c11Phy}
	\end{equation}
	the following holds:
	\begin{align}
		&-\lambda t_{u^{\prime\prime}}^{(k)}\log(r_{u^{\prime\prime}})+(1-\lambda) w_{u^{\prime\prime}}^{(k)} \nonumber\\&< -\lambda t_{u^{\prime}}^{(k-1)}\log(r_{u^{\prime}})+(1-\lambda) w_{u^{\prime}}^{(k-1)}.
		\label{c9Phy}
	\end{align}
	If $u^{\prime\prime}$ exists, we remove a time slot from that user and assign it to user $u^\prime$. Therefore,
	\vspace{-0.3cm}
	\begin{equation}
		t_{u^{\prime}}^{(k)}=t_{u^{\prime}}^{(k-1)}+1
		%t_{u^{\prime}}^{(k)}=t_{u^{\prime\prime}}^{(k-1)}+1
		\label{c12Phy}
	\end{equation}
	and \vspace{-0.3cm}
	\begin{equation}
		w_{u^{\prime}}^{(k)}=R_{u^{\prime}}^{e}-\frac{t_{u^{\prime}}^{(k)}}{T_B}\cdot r_{u^{\prime}}.
		\label{c13Phy}
	\end{equation}
	
	We keep iterating, finding both $u^\prime$, and $u^{\prime\prime}$, and pass one time slot from user $u^{\prime\prime}$ to user $u^{\prime}$ until we reach the point where there is no such user $u^{\prime\prime}$. This would be the convergence point and the corresponding time allocation is the final solution to the proposed optimization problem. We clearly see that the algorithm's complexity is $O(U)$. Intuitively, the algorithm tries to push the user with the minimum rate to achieve a higher rate but not at the cost of pushing the rate of another user below the current minimum. Therefore, at each iteration, the minimum rate is improved.
	  
	\begin{table}[t]
		\vspace{.05in}
		\caption{Algorithm 1: The Proposed Algorithm to Solve the Physical Layer Subproblem } % title of Table
		%\centering % used for centering table
		\begin{tabular}{c l } % centered columns (4 columns)
			\hline\hline %inserts double horizontal lines
			%Case & Method\#1 \\ [0.5ex] % inserts table
			%heading
			%\hline % inserts single horizontal line
			&   \\
			& \textbf{Input}: $T, T_B, \lambda, U, N, B, P, f_{u,n}^t, \forall u,n, R_u^e\forall u$\\ 
			&$B$, $P_{max}$, $H_k\forall k$, $N_0$, $\gamma$ \\ % inserting body of the table
			& \textbf{Output}: $t_u,\forall u \in\{1, \dots, U\}$ \\
			1 & \textbf{Initialize} : $k=0, K=$max no of iterations\\
			2 & assign random number of time slots to users such that:\\
			& $\sum_{u=1}^Ut_u=T$ \\
			3 & $k \leftarrow k+1$\\
			4 & \textbf{While}  $k<K$ \\
			5 & \,\,\, Find $u_\prime^{(k)}$ according to \eqref{c8Phy}\\
			6 & \,\,\,\ Find $u_{\prime\prime}^{(k)}$ such that when \eqref{c10Phy}-\eqref{c11Phy} performed, \eqref{c9Phy} holds \\
			7 & \,\,\,\, \textbf{if}  user $u_{\prime\prime}^{(k)}$ is not exist\\
			8 & \,\,\,\,\,\,\,  break\\
			9  & \,\,\,\, Update $t_{u^\prime}^{(k)}$ and $w_{u^\prime}^{(k)}$ according to  \eqref{c12Phy}-\eqref{c13Phy}\\
			10 & \,\,\,\, $k \leftarrow k+1$\\
			11   & \textbf{endWhile}\\
			&  \\ [1ex] % [1ex] adds vertical space
			\hline %inserts single line
		\end{tabular}
		\label{table:bbm} % is used to refer this table in the text
	\end{table}  
	\subsection{Optimality Condition of the Proposed Algorithm}
	
	In this subsection, we show that when $\lambda=0$, the proposed algorithm solves the physical layer scheduling problem optimally.  Intuitively,  when $\lambda=0$, the objective of the problem becomes minimizing the AFER violation. Mathematically,  When $\lambda=0$, the objective function reduces to:
	\begin{equation}
		\text{\textbf{Minimize} }  {\text{Max}}({\cal W})
		\label{equ:mainObjCase1}
	\end{equation}
	Moreover, equation \eqref{c8Phy} reduces to:
	\begin{equation}
		u^\prime=\underset{u}{\text {argmax}}(w_{u})
		\label{c8PhyCase1}
	\end{equation}
	and  equation \eqref{c9Phy} reduces to:
	\begin{equation}
		w_{u^{\prime\prime}}^{(k)} < w_{u^{\prime}}^{(k-1)}
		\label{c9PhyCase1}
	\end{equation}
	Hence, finding user $u^{\prime\prime}$ reduces to finding a user in which when one time slot is taken from her, the maximum rate violation at iteration $k$ is lower than its value at time slot $k-1$.
	
	{\bf {Lemma 1: }} {\it Using Algorithm 1, if $F^{(k-1)}$ is the value of the objective function at iteration $k-1$, and $F^{(k)}$ is the value of the objective function at iteration $k$, then the following holds true when $\lambda = 0$:
		\begin{equation}
			F^{(k)} \leq F^{(k-1)}
			\label{lemma1Eq}
		\end{equation}
		In other words, the objective function is non increasing at every iteration.}
	
	{\bf {Proof: }} At iteration \smash{$k\!\!-\!\!1$}, we know that \smash{$w_{u^{\prime}}^{(k\!-\!1)}\!\!\!=\!\textbf{Max}({\cal W}^{(k\!-\!1)})$}. Moreover,  the only change from iteration $k-1$ to iteration $k$ is taking a time slot from user $u^{\prime\prime}$ and giving it to user $u^{\prime}$ such that $w_{u^{\prime\prime}}^{(k)} < w_{u^{\prime}}^{(k-1)}$. Since  $w_{u^{\prime\prime}}^{(k)} < w_{u^{\prime}}^{(k-1)}$, $\textbf{Max}({\cal W}^{(k)})$ will never be greater than $\textbf{Max}({\cal W}^{(k-1)})$ and that concludes the proof.

	{\bf {Theorem 1: }} {\it Algorithm 1 Converges to the optimal solution of the optimization problem \eqref{equ:mainObjPhy}-\eqref{phyC4} when $\lambda = 0$.}
	
	{\bf {Proof: }} Lemma 1 proves that the objective function is non increasing over iterations. Moreover, the proposed algorithm's task at every iteration is to take a time slot from user $u^{\prime\prime}$ if it exists and give it to user $u^{\prime}$ such that $w_{u^{\prime\prime}}^{(k)} < w_{u^{\prime}}^{(k-1)}$. Therefore, at the convergence point, there is no existence of user $u^{\prime\prime}$, and  the objective function is at its minimum. Otherwise, there will be a user $u^{\prime\prime}$ such that $w_{u^{\prime\prime}}^{(k)} < w_{u^{\prime}}^{(k-1)}$ and the algorithm would have selected it to further reduce the objective function's value.
	
	%\vspace{-0.1cm}
\section{Proposed Algorithm to Solve The Application Layer subproblem}
	\label{algo_appLayer}
The application layer subproblem  is non-convex with non-causal knowledge of the user's FoV. Moreover, the practical VBR encoding of video tiles makes the quality decisions more difficult. Therefore, we propose an efficient algorithm based on ML. In particular, we consider a distributed reinforcement learning approach to find the quality decisions based on real observations. 

Our approach, described in Fig~\ref{fig:act}, considers a reward that captures the dependency of each agent's action on the other agents' actions. Before we describe our model, we first give a brief introduction to RL. In RL, there are two components known as agent and environment in which the agent interacts with the environment, performs some action and receives a reward. The RL observes a state $s_{k}$, and performs an action $a_{k}$. After performing the action, the state of the system transitions to $s_{k+1}$, and the agent receives a reward $R_k$. In the learning phase, the agent learns from the rewards it receives at every (state, action) pair $(s_k,a_k$) how good is being at that state $s_k$, and what is the best action $a_k$ to take. The objective of learning is to maximize the expected cumulative discounted reward defined as $E\{\sum_{k=0}^{\infty}\sum_{t=0}^{\infty}\gamma^t\eta_{k+t}\}$. 
	
However, considering one RL agent in our problem to make the quality decisions of all tiles will lead to exponential number of actions that the RL needs to choose from for each tile. To avoid high computation complexity, we propose an approach that uses RL agent for each tile. Therefore, each user will run a number of RL agents that is equal to the number of chunk tiles.  At the download of each chunk, the RL agents of user $u$ are run sequentially (one after the other), i.e., RL agent $j$ observes a state $s_{c,j}$ which includes the remaining bandwidth after all tiles $1$ to $j-1$ have decided their download rate (action) and performs an action $a_{c,j}$, where $c$ and $j$ are the chunk and tile indices, respectively. After performing the action, the state transitions to $s_{c+1,j}$, and the RL agent receives a discounted reward. The discounted reward does not only consider the time (as in conventional RL) but also  the space (other agents), since the action of the $j$-th RL agent does not only affect the future actions of this agent, but also the actions of all agents $j+1$ to $J$. Moreover, the reward is delayed since it depends on the FoV of the user at the playback of the chunk.  The time discounted, space aware, and delayed reward we consider is, $E\{\sum_{c=1}^{C}\sum_{j=1}^J\sum_{t=1}^{\infty}\gamma^{t-1}\eta_{c+t}^j\}$, where $\eta_c^j$ is defined in~\eqref{equ:reward}. This choice considers the effect of the $j$-th agent's action on the other tiles and ensures a long term system reward of the current action. 

As shown in Fig. \ref{fig:act}, the quality decision of each tile in the next chunk to download is obtained from training a neural network. Each tile's RL agent observes a set of metrics including the probability that tile $j$ is in the FoV of the chunk to download, as well as the available user's bandwidth that is obtained by solving the physical layer problem. The agent feeds these parameters to the neural network, which outputs the action. The action is defined by the quality of this tile in the next chunk to download. The total reward is then observed and fed back to the agent to train its neural network model. Our reward function, state, and action spaces are defined as follows:
	
	\begin{itemize}
		\item {\bf Reward Function}: The reward function at the download of tile $j$ in chunk $c$, $\eta_c^j$ is defined as: 
		\begin{equation}
			\eta_c^j= \sum_{\substack{k=j,\\ j\in\text{ Fov}}}^{J}\sum_{q=0}^{Q}\beta^{q}I_{u,c}^{k,q}
			\label{equ:reward}
		\end{equation}
		
		\item {\bf State}: The state at the download of tile $j$ of chunk $c$ is:
		\begin{itemize}
			\item $G_u^j$: the remaining size that user $u$ can download given the rate that the user can achieve, the total number of time slots assigned to this user, and how many time slots are left to download tiles $j$ to $J$. Note that $G_u^1=\hat{R}_u\cdot t_u$ since when making the decision of the first tile in the chunk, the remaining size is exactly the total size that user $u$ can download.
			\item $\textbf{Pr}_{c,j}\in[0,1]$: the probability that tile $j$ is in the FoV of chunk $c$, which can be calculated from the FoV of other users who watched the same video in the past. 
			\item ${\textbf F}_{c^\prime,j} \in\{0,1\}$: an indicator variable that is equal to $1$ if tile $j$ is in the FoV of the chunk that is currently being played ($c^\prime$), and $0$ otherwise. This indicator is found by observing the current FoV of the user.
			\item $\zeta$: a vector of  the available sizes of the next tile to download. The sizes are corresponding to the available quality levels.
		\end{itemize}
		\item {\bf Action}: The action is represented by a probability vector of length $N$ such that if the $q$-th element is the maximum, the tile $c$ will be downloaded at the $q$-th quality level.
	\end{itemize}
	\if0 
	\begin{figure}
		\centering
		\includegraphics[trim=1in 0.4in 0.9in 0in, clip, width=.5\textwidth]{./Actor_critic_AoI.pdf}	
		\caption{The Actor-Critic Method that is used in our Proposed Scheduling Algorithm}
		\label{fig:act}
	\end{figure}
	\fi 

        \begin{figure}
		\centering
		%trim=1in 0.4in 0.9in 0in, clip, width=\textwidth
		%scale=0.8
		\includegraphics[trim=0in 0in 0in 0in, clip,width=0.5\textwidth]{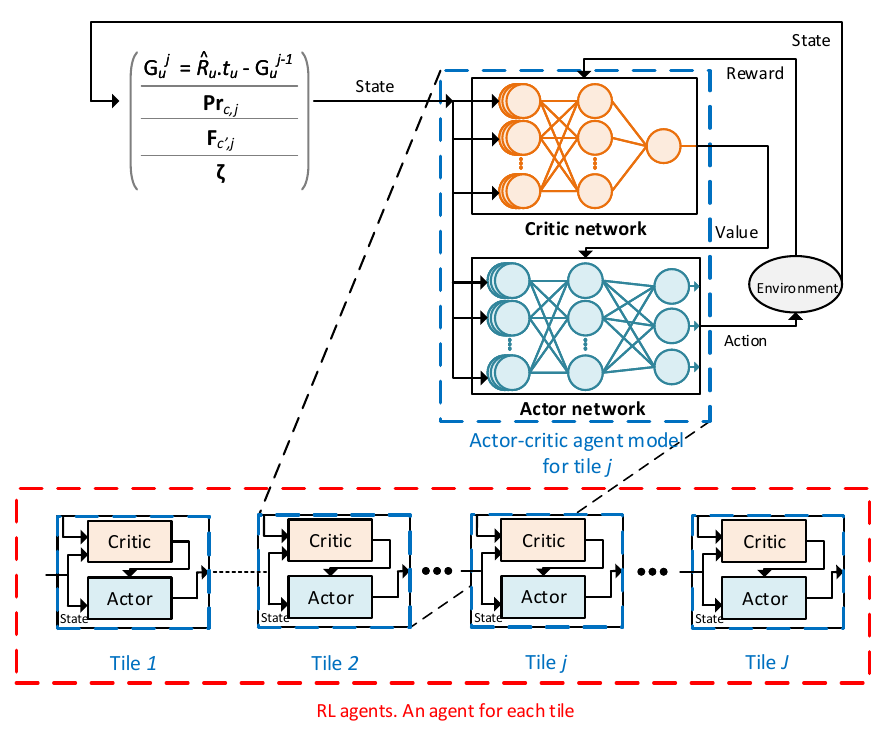}	
		\caption{The proposed reinforcement learning based scheduling algorithm for solving the application layer problem.}
		\label{fig:act}
	\end{figure}
 
	\if0
	\begin{table}[t]
		\vspace{.05in}
		\caption{Proposed Algorithm to Solve Application Layer Problem Per Chunk} 
		\begin{tabular}{c l } 
			&   \\
			& \textbf{Input}: $\hat{R}_u^(t), Pr_{c,t}, F_{c^\prime,t}\forall t, \zeta$\\ 
			&At the download of each chunk:
			& Perform action of each 
			1 & \textbf{Initialize} : $k=0, K=$max no of iterations\\
			2 & assign random number of time slots to users such that:\\
			& $\sum_u=1^Ut_u=T$ \\
			3 & $k \leftarrow k+1$\\
			4 & \textbf{While}  $k<K$ \\
			5 & \,\,\, Find $u_\prime^{(k)}$ according to \eqref{}\\
			6 & \,\,\,\ Find $u_{\prime\prime}^{(k)}$ such that when \eqref{}-\eqref{} performed, \eqref{} holds \\
			7 & \,\,\,\, \textbf{if}  user $u_{\prime\prime}^{(k)}$ is not exist\\
			8 & \,\,\,\,\,\,\,  break\\
			9  & \,\,\,\, Update $t_{u^\prime}^{(k)}$ and $w_{u^\prime}^{(k)}$ according to  \eqref{}-\eqref{}\\
			10 & \,\,\,\, $k \leftarrow k+1$\\
			11   & \textbf{endWhile}\\
			&  \\ [1ex] % [1ex] adds vertical space
			\hline %inserts single line
		\end{tabular}
		\label{table:bbm} % is used to refer this table in the text
	\end{table}
	\fi
	
To train our RL based system, we use the Asynchronous Advantage Actor-Critic (A3C) framework \cite{mnih2016asynchronous} which is the state-of-art actor-critic algorithm. A3C involves training two neural networks. The agent selects an action based on a policy defined as a probability over actions $\pi : \pi(s_{c,j},a_{c,j}) \leftarrow [0,1]$. The policy is a function of a \emph{policy parameter} $\theta$. Therefore, for each choice of $\theta$, we have a parametrized policy $\pi_\theta(s_{c,j},a_{c,j})$.
	
Policy gradient method~\cite{sutton2000policy} is used to train the actor-critic policy. Here, we describe the main steps of the algorithm.
Policy gradient methods estimate the gradient of the expected cumulative discounted reward with respect to the policy parameters $\theta$, which is computed as:
	\begin{equation}
		\nabla\! E\{\!\sum_{c=1}^{C}\!\sum_{k=j}^{J}\!\sum_{t=1}^{\infty}\!\!\gamma^{t-1}\eta_{c+t}^j\}\!\!=\!\!E_{\pi_{\theta}}\!\{\nabla_{\!\theta}\!\log \pi_\theta(s,\!a)A^{\pi_\theta}\!(s,\!a)\!\}
	\end{equation}
where $A^{\pi_\theta}(s,a)\}$ is the advantage function~\cite{mnih2016asynchronous}. The agent uses an empirically computed advantage as an unbiased estimate of $A^{\pi_\theta}(s_{c,t},a_{c,t})\}$ by sampling a trajectory of scheduling decisions. The update of the actor network parameter $\theta$ follows the policy gradient, which is defined as follows:
	\begin{equation}
		\theta \leftarrow \theta + \alpha \sum_{c=1}^{C}\nabla_\theta \log \pi_\theta(s_{c,t},a_{c,t})A^{\pi_\theta}(s_{c,t},a_{c,t})
		\label{policyUpdate}
	\end{equation}
	where $\alpha$ is the learning rate. To compute the advantage $A(s_{c,t},a_{c,t})$ for a given action, we need to estimate the value function of the current state ${\cal{V}^{\pi_\theta}}(s)$ which is the expected total reward starting at state $s$ and following the policy $\pi_\theta$. Estimating the value function from the empirically observed rewards is the task of the critic network. To train the critic network, we follow the standard temporal difference method~\cite{sutton2011reinforcement}. In particular, the update of $\theta_v$ follows the following equation:
	\begin{equation}
		\theta_v \leftarrow \theta_v\!+\!\alpha^\prime\!\sum_{c=1}^{C}\big(\eta_c\!+\! \gamma V^{\pi_\theta}(s_{c+1},\theta_v)\!-\!V^{\pi_\theta}(s_c,\theta_v)\big)^2
		\label{valueUpdate}
	\end{equation}
	where $\alpha^\prime$ is the learning rate of the critic, and $V^{\pi_\theta}$ is the estimate of $\cal{V}^{\pi_\theta}$. Therefore, for a given ($s_{c,t},a_{c,t},\eta_c, s_{c+1,t}$), the advantage function, $A(s_{c,t},a_{c,t})$ is estimated as $\eta_c+\gamma V^{\pi_{\theta}}(s_{c+1,t},\theta_{v})-V^{\pi_{\theta}}(s_{c,t},\theta_{v})$.
	
Finally, we note that the critic is only used at the training phase to help the actor converge to the optimal policy.

	%\vspace{-0.1cm}
	\section{Performance Evaluation}
	\label{sec:sim}
	 
	In this section, we numerically evaluate the performance of our proposed approach. To
	better illustrate the impact of its cross layer design, we will compare it against two baseline
	schemes that do not jointly consider the physical and application layer problems. The
	considered baselines observe the following rate adaptation techniques at the application
	layer once the tiles comprised in each chunk have been sorted according to their probability to be in the FoV:
	
	\begin{itemize}
		\item Quality-prioritized Selection (\texttt{QPS}): 
		Starts by assigning the highest possible quality to the most probable tile, it will then proceed to consider the quality for the second most probable tile, and so on. This algorithm follows a vertical strategy in which it prioritizes pushing the current tile to the highest quality before making the decision about the next tile in descending probability order.
		\item FoV-prioritized Selection (\texttt{FPS}): Initially assigns the base quality to all tiles sequentially. If after granting this quality there is still available bandwidth, it resumes the process by increasing sequentially the quality level to the second worst for all tiles. This process is repeated until the highest quality level is reached or there is no available bandwidth anymore. To avoid wasting bandwidth in tiles with very low FoV probability, the algorithm can also come to an end at a predefined threshold, such that tiles
		with lower qualities are not scheduled. As opposed to QPS, this algorithm follows a horizontal strategy. i.e., the algorithm prioritizes pushing more tiles to the next quality level to avoid missing any tile within the FoV.
	\end{itemize}
	
For both baseline schemes, it is assumed that bandwidth prediction is leveraged to make their quality decisions. Hence, we consider a moving average procedure as it provides a good simplicity vs. performance trade-off. Specifically, the rate achieved over the last $100$ time slots is considered to predict the future bandwidth. 
	
	To highlight the impact of each of the two subproblems, the rate allocation problem (the physical layer problem) is solved for both schemes in two different ways, either using 
	a proportional fairness scheduler or our proposed solution, and labeled $\ast\texttt{-PF}$ and $\ast\texttt{-PR}$, respectively. Therefore, a total of four baselines, denoted  $\texttt{QPS-PF}$, $\texttt{FPS-PF}$, $\texttt{QPS-PR}$, and $\texttt{FPS-PR}$, in addition to our proposed cross layer approach, referred to as $\texttt{PROPOSED}$ are compared. Note that, after the rates are allocated, the actual quality levels are obtained using the application layer logic.
	
	To conduct our  performance comparison, we consider a realistic dataset of users watching 360 degrees videos \cite{LOVRDataset17}. The dataset consists of $50$ users, watching $10$ YouTube 360 degree videos. Each video is sampled at $30$ frames-per-second and the data for $1800$ frames are recorded, which correspond to $1$ minute of playback. Each frame consists of $200$ tiles, which together form a $10\text{\texttimes}20$ grid in the equirectangular projection. Due to the high number of tiles, we reduce the tile grid dimension to a $5\text{\texttimes}10$
	grid ($50$ tiles). Accordingly, each tile in the reduced grid is equivalent to $4$ tiles ($2\text{\texttimes}2$ grid) in the original representation, and is considered in the user's FoV if any of the $4$ tiles overlapped with the FoV. 
	
The architecture of each RL agent in our distributed RL framework is based on a modified version of the Pensieve deep RL architecture \cite{mao2017neural}. Each RL agent comprises both the actor and the critic network, where the $\zeta$ vector is fed to a 1D convolutional neural network (CNN) layer with $128$ filters, each with $4\text{\texttimes}4$ size and stride $1$. The output of the CNN layers as well as all other inputs are then aggregated in a hidden layer with $128$ neurons and a rectified linear unit (ReLU) activation layer. Finally, in the actor
network, the output is passed to a softmax layer with $5$ outputs, to construct a probability distribution corresponding to the quality selection policy. The discounted reward is enforced by selecting $\gamma=0.9$, while the actor-critic learning rates are both set to $0.001$. The Network is implemented and trained using the TensorFlow library.
	
We assume a network with $5$ users watching $5$ different videos, and whose distances from their serving SBSs are selected in the range between $15$ m and $35$ m.  Simulations are run for a total of $50,000$ iterations. The channel  between a user and an SBS is calculated taking into account both pathloss and Rayleigh fading, with a pathloss exponent of $2$. The downlink transmit power is $45$ dBm and the mmWave bandwidth is $1$ GHz, whereas the noise power spectral density is assumed to be $10^{-9}$ Watt/Hz and the beam gain is set as $f_{u,n}=10$. Each chunk transmission period starts by a beam alignment period that occupies $5\%$ of the chunk download time, after which we assume that the SBS and user are perfectly aligned for the chunk transmission period.
	
FoV traces are obtained from real traces collected for users watching $5$ videos selected from the dataset in \cite{LOVRDataset17}. We use $70\%$ of these traces ($35$ users) to calculate the probability of a tile being in the FoV, while the remaining $30\%$ are used in our simulations. Furthermore, we use the same data to simulate the baseline schemes where the \texttt{FPS} variants are implemented with a probability threshold $p_{\textrm{th}}=0.01$.
	
	\begin{figure}[t]
		\centering\includegraphics[trim=3.1in 0.5in 3.1in 0.5in,width=3.7cm]{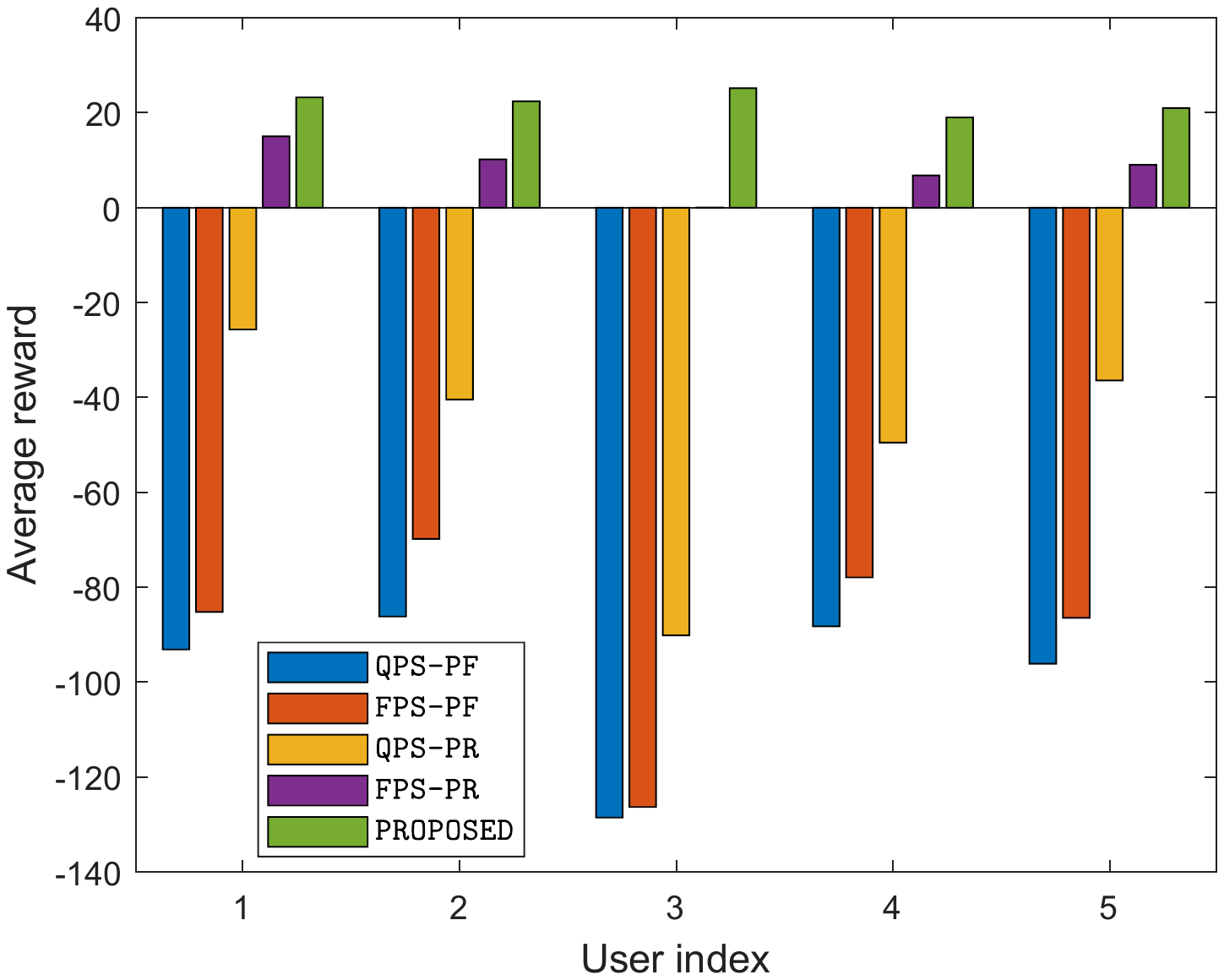}
		
		\caption{The average reward of the different schemes for $5$ users watching $5$ different 360 degree
			videos.}
		\label{fig:avg_reward}
		 \vspace{-0.5cm}
	\end{figure}

	\begin{figure*}
		\centering
		\subfloat[]{\includegraphics[trim=1in 0.4in 0.9in 0in,width=6cm]{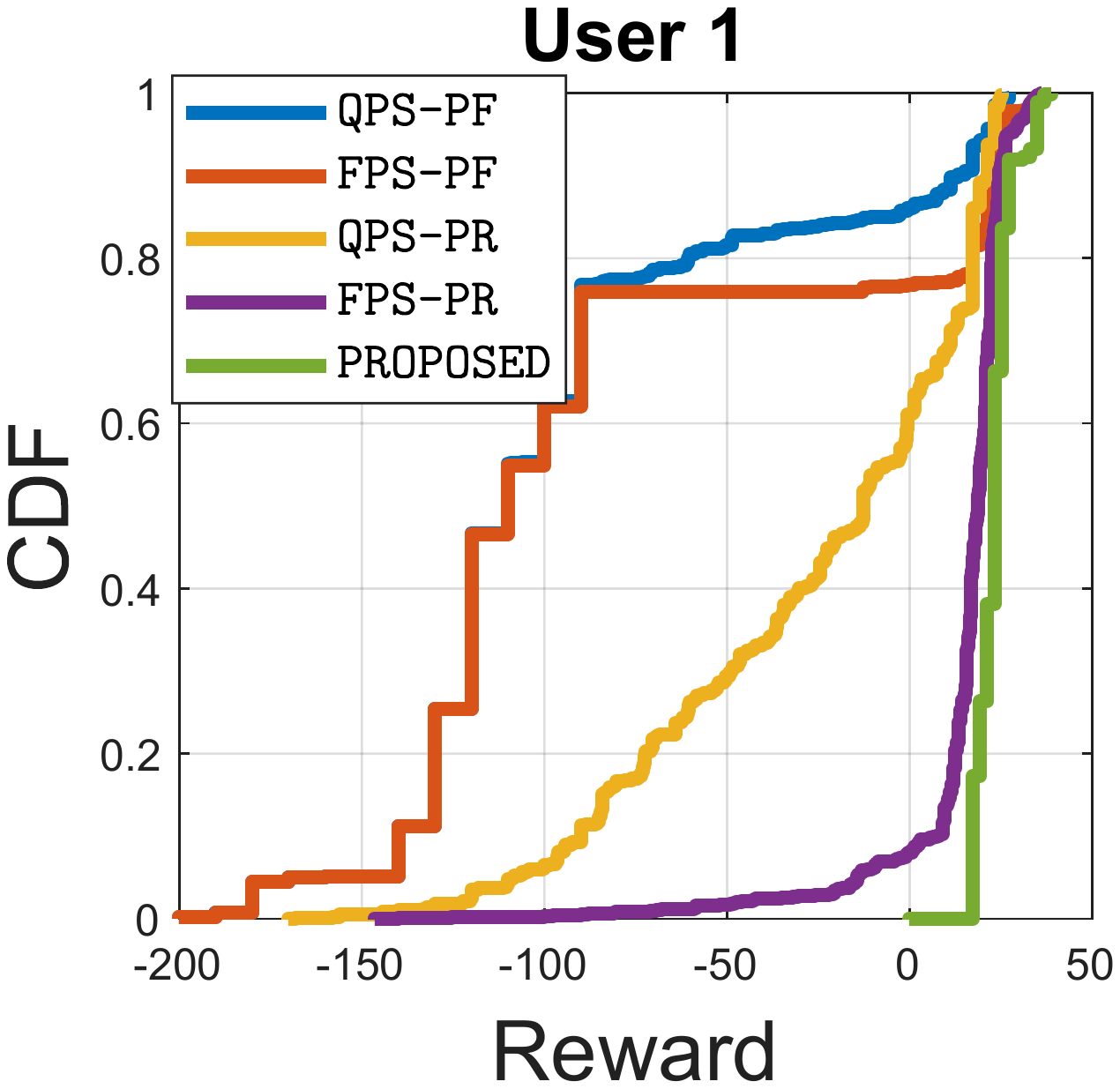}\hspace{-0.35cm}
		}
		\subfloat[]{\includegraphics[trim=1in 0.4in 0.9in 0in,width=6cm]{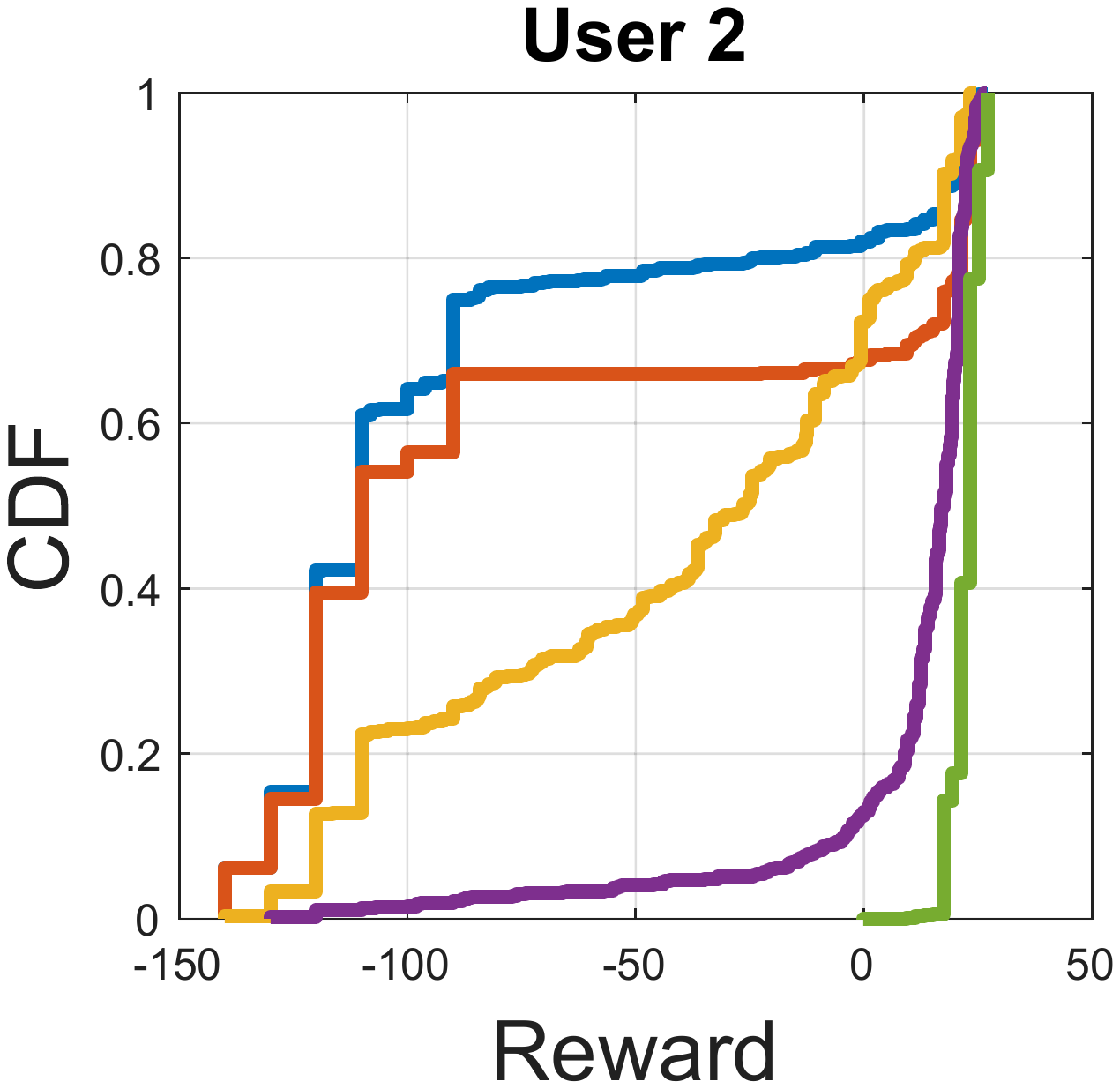}\hspace{-0.35cm}
		}
		\subfloat[]{\includegraphics[trim=1in 0.4in 0.9in 0in,width=6cm]{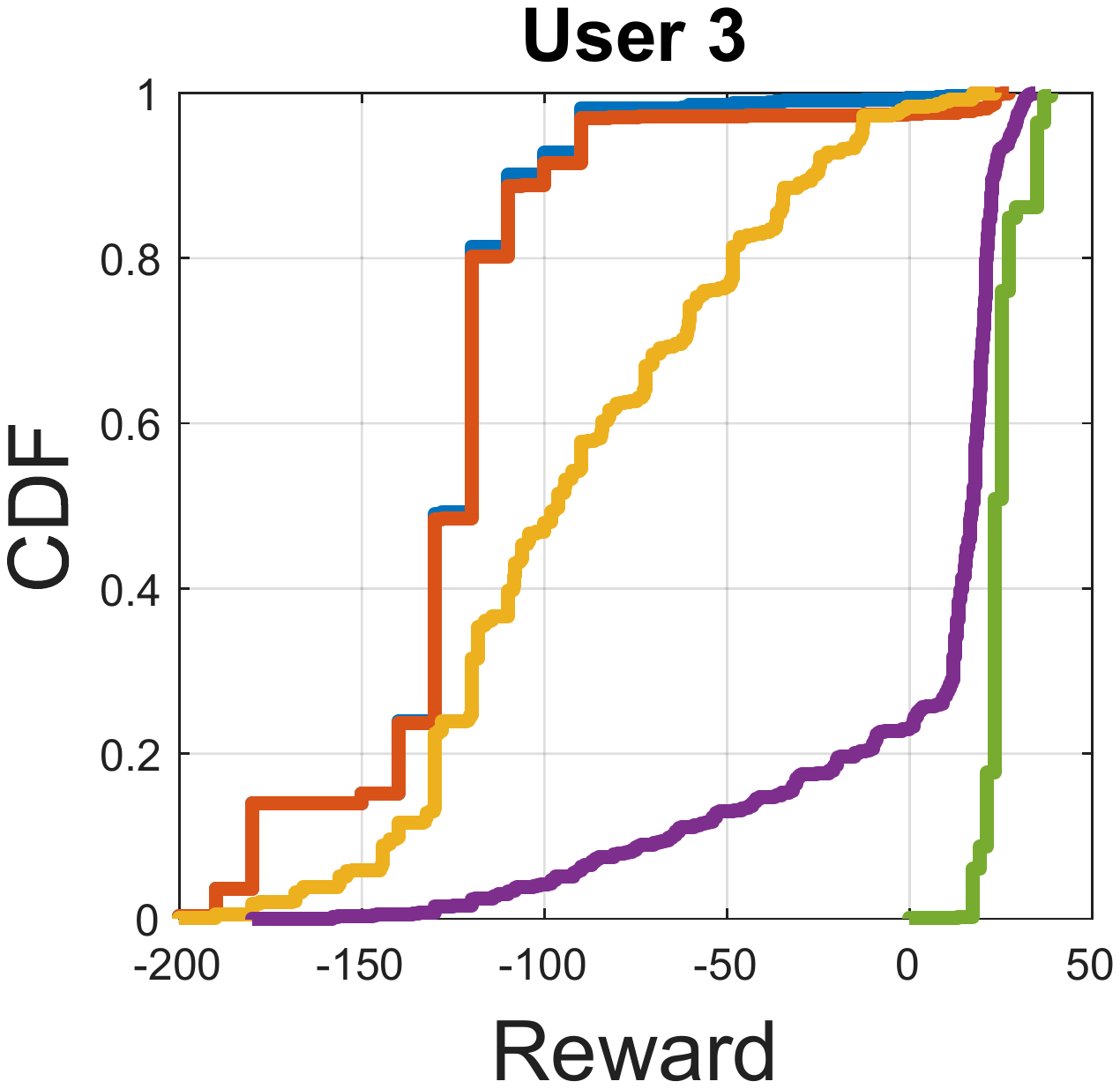}\hspace{-0.35cm}
		}
		\\
		\subfloat[]{\includegraphics[trim=1in 0.4in 0.9in 0in,width=6cm]{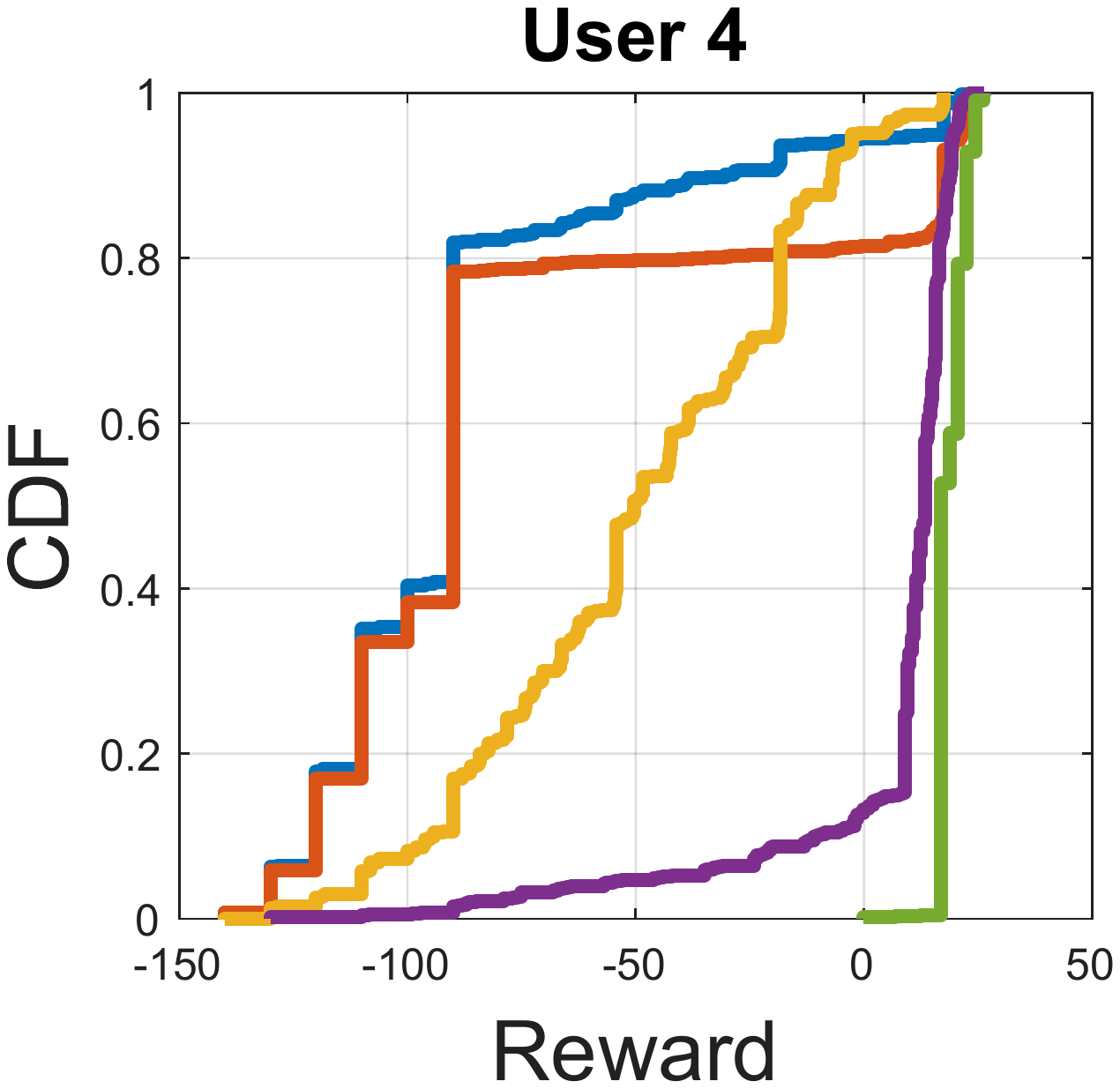}\hspace{-0.35cm}
		}
		\subfloat[]{\includegraphics[trim=1in 0.4in 0.9in 0in,width=6cm]{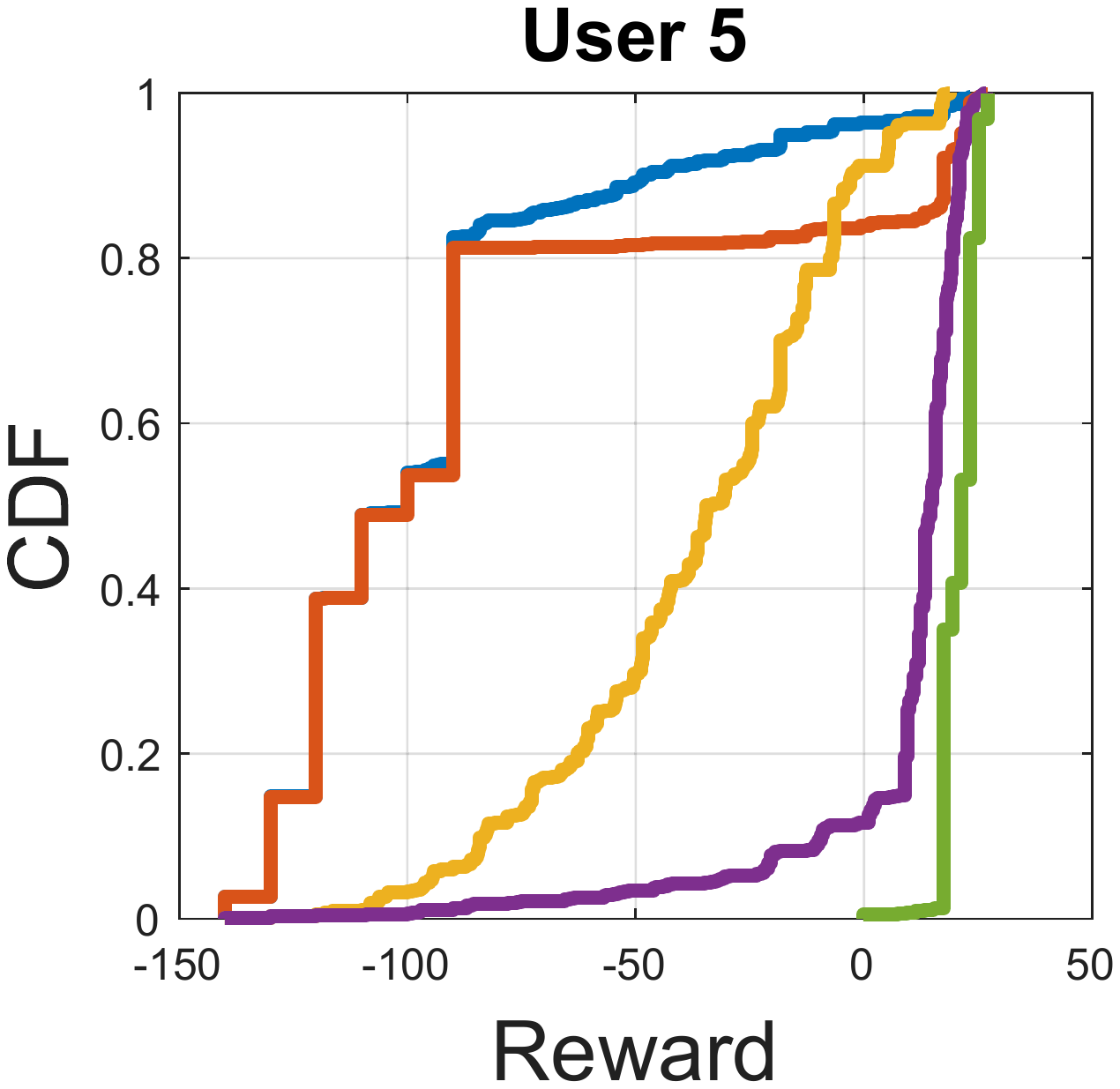} 
		}
		\caption{The CDF of the reward function for $5$ users watching $5$ different 360 degree videos.}
		\label{fig:CDFs}
	\end{figure*}

	In Fig. \ref{fig:avg_reward}, we plot the average reward of the different schemes when the $5$ users are watching $5$ different 360 degree videos. Therein, we can see that the schemes with the proportional fairness scheduler fail to achieve reasonable rates, i.e. sufficient for a basic streaming quality. 
	A high miss rate of FoV tiles results in negative reward values in both  $\texttt{QPS-PF}$ and $\texttt{FPS-PF}$, which are considerably less than their values in $\texttt{QPS-PR}$ and $\texttt{FPS-PR}$. Note that $\texttt{QPS-PR}$ and $\texttt{FPS-PR}$ use the proposed physical layer approaches. 
	Moreover, comparing $\texttt{QPS-PR}$ and $\texttt{FPS-PR}$, it can be shown that $\texttt{QPS-PR}$ achieves a lower average reward. The decrease is due to the high penalty paid for missing
	a high number of tiles in the FoV. This is a consequence of the nature of $\texttt{QPS}$ schemes, that prioritize assigning the highest quality to the tiles with high FoV probability, which result in failing to download the tiles with lower FoV probability. Both our $\texttt{PROPOSED}$ and $\texttt{FPS-PR}$ schemes achieve better performance in terms of the average reward. Indeed, our $\texttt{PROPOSED}$ scheme achieves significantly higher gains even in challenging situations (e.g., user $3$ watching video $3$).
	
	In Fig. \ref{fig:CDFs}, we plot the cumulative distribution function (CDF) of the reward achieved by the different schemes while running on each of the five users. Here, the proposed scheme is shown to achieve significantly higher rewards most of the time with all of the videos. In comparison, $\ast\texttt{-PR}$ schemes achieve higher performance than the
	the $\ast\texttt{-PF}$ schemes. The latter is due to the ability of $\ast\texttt{-PR}$ to be \emph{QoE aware}, pushing all users to get, at least, the minimum rate that guarantees downloading all FoV tiles; even if that means getting them at the lowest quality level. In the application layer, $\texttt{QPS}$ schemes are shown to achieve lower rewards since they focus on increasing the quality of the tiles with high FoV probability which may lead to incurring in high penalties for missing tiles in the FoV. In this regard $\texttt{FPS}$ baselines achieve a better performance since they prioritize assigning possibly low but non-zero quality levels to all the tiles with probability to be in the FoV above the threshold $p_{\textrm{th}}$. Despite the good average performance, \texttt{FPS} schemes, however, fail to perform well in terms of low-percentile values since their probability-based prioritization fails to capture the outlier cases. In contrast, our DRL-based proposed approach achieves significantly higher rewards by leveraging the exploration-exploitation nature of RL.

	\begin{figure}[h]
		\centering\includegraphics[trim=3.1in 0.4in 3.1in 0.5in,width=3.7cm]{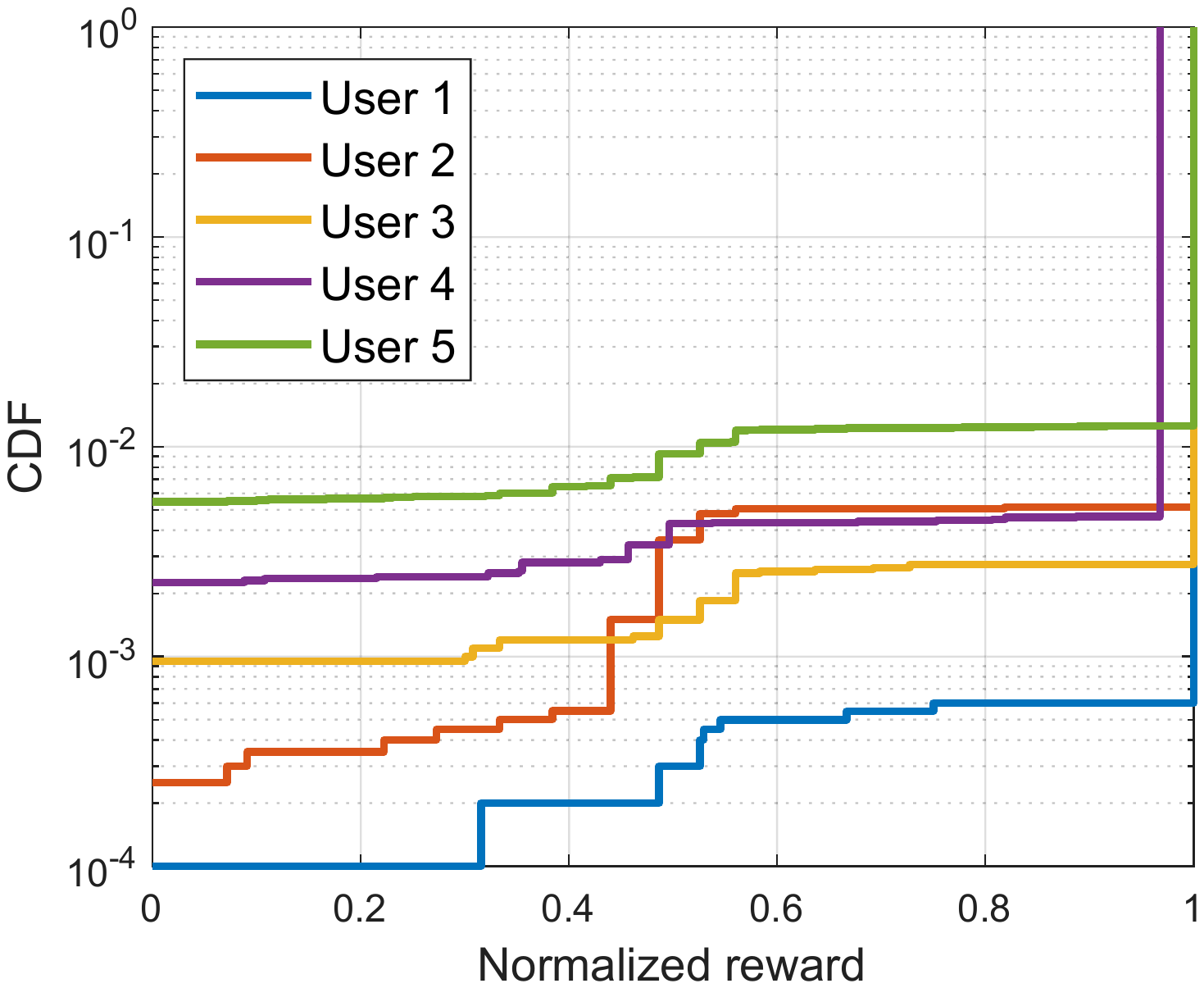}
		
		\caption{The CDF of the normalized reward of the $\texttt{PROPOSED}$ scheme with respect to the maximum achievable reward per frame.}
		\label{fig:norm_reward}
		
	\end{figure}
	Finally, a closer look at the performance of the $\texttt{PROPOSED}$ scheme is provided in Fig. \ref{fig:norm_reward}. In particular, we investigate the normalized reward performance with respect to the maximum reward a system can get. The maximum reward is calculated assuming a system that has full knowledge of a user's FoV tiles, as well as full capacity to deliver those tiles in the highest quality. The quality for each tile in the FoV can then be expressed as $\sum_{q=0}^{Q-1}\beta^{q}$.  Fig. \ref{fig:norm_reward} shows the CDF of the normalized CDF for the $5$ users using our $\texttt{PROPOSED}$ scheme. It exposes that our cross layer approach matches the maximum achievable reward with ultra-high probability for four users, whereas a single user slightly deviates from this maximum achievable reward. This result clearly demonstrates that our cross-layer approach is a powerful method to maximize the users' QoE in 360 degree video streaming scenarios.

	% \vspace{-0.1cm} 
	\section{Conclusion}\label{sec:conc}
 
	In this paper, we have proposed a cross layer optimization approach for multi-user 360 degree video streaming. The proposed framework maximizes the available rate to each user and allows each user to efficiently utilize the available rate to maximize its QoE. A QoE metric that maintains a tradeoff between maximizing each user's QoE, and ensuring fairness among users has been considered. We have shown that the main problem can be decoupled into two subproblems: (i) the physical layer subproblem whose objective is to find the download rate of each user, and (ii) the application layer subproblem whose objective is to use that rate to find a quality decision per tile such that the user's QoE is maximized. We have shown that the physical layer subproblem can be solved optimally with low complexity. Moreover, we have proposed a reinforcement learning based approach to solve the application layer subproblem. Our conducted experiments have revealed the robustness of our scheme and demonstrated its significant performance improvement compared to several 360 degree streaming algorithms.
	 
	  \vspace{-0.3cm}
 \bibliographystyle{IEEEtran} %
 
 	\bibliography{IEEEabrv,refs,ref2}%,bib,multipath_feng}

\end{document}